\documentclass[a4]{article}
\usepackage{multirow}
\usepackage{graphicx}

\usepackage{authblk}
\usepackage{amsfonts}
\usepackage{amsmath}
\usepackage{url}
\newcommand{\argmin}{\operatornamewithlimits{argmin}}

\title{Mapping Tractography Across Subjects}

\author{Thien Bao Nguyen}
\author{Emanuele Olivetti}
\author{Paolo Avesani
  \thanks{The research was funded by the Autonomous Province of
    Trento, Call ``Grandi Progetti 2012'', project ``Characterizing and
    improving brain mechanisms of attention - ATTEND''.  }}

\affil{NeuroInformatics Laboratory (NILab), \\
  Bruno Kessler Foundation, Trento, Italy}

\affil{Center for Mind and Brain Sciences (CIMeC), \\
  University of Trento, Italy}

\affil{\tt{http://nilab.fbk.eu}}

\date{}

\begin{document}

\maketitle

\begin{abstract}
  Diffusion magnetic resonance imaging (dMRI) and tractography provide
  means to study the anatomical structures within the white matter of
  the brain. When studying tractography data across subjects, it is
  usually necessary to align, i.e. to register, tractographies
  together. This registration step is most often performed by applying
  the transformation resulting from the registration of other
  volumetric images (T1, FA). In contrast with registration methods
  that \emph{transform} tractographies, in this work, we try to find
  which streamline in one tractography correspond to which streamline
  in the other tractography, without any transformation. In other
  words, we try to find a \emph{mapping} between the
  tractographies. We propose a graph-based solution for the
  tractography mapping problem and we explain similarities and
  differences with the related well-known graph matching
  problem. Specifically, we define a loss function based on the
  pairwise streamline distance and reformulate the mapping problem as
  combinatorial optimization of that loss function. We show
  preliminary promising results where we compare the proposed method,
  implemented with simulated annealing, against a standard
  registration techniques in a task of segmentation of the
  corticospinal tract.
\end{abstract}

\section{Introduction}
\label{sec:introduction}
Diffusion magnetic resonance imaging (dMRI)~\cite{basser1994mr} is a
modality that provides non-invasive images of the white matter of the
brain. DMRI measures the local the diffusion process of the water
molecules in each voxel. That process represents structural
information of neuronal axons. From dMRI data, tracking
algorithms~\cite{mori2002fiber,zhang2008identifying} allow to
reconstruct the $3D$ pathways of axons within the white matter of the
brain as a set of streamlines, called tractography. A
\emph{streamline} is a 3D polyline representing thousands of neuronal
axons in that region of the brain, and a \emph{tractography} is a
large set streamlines, usually $\approx 3 \times 10^5$.


Current neuroscientific analyses of white matter tractography data are
limited to qualitative intra-subject comparisons. Thus, it is quite
difficult to use the information for direct inter-subject
comparisons~\cite{golding2011comparison,bazin2011direct}. This leads
to the need of initial alignment, or registration, of tractographies
via some methods before doing further study. Registration is most
often performed by applying the transformation resulting from the
registration of other images, such as $T1$ or fractional anisotropy
(FA), to
tractography~\cite{goodlett2009group,golding2011comparison,wang2011dti}. Recently,
\cite{odonnell2012unbiased} proposed group-wise registration using the
trajectory data of the streamlines. The idea to work on deterministic
tractography rather than other images is quite innovative. And, it may
be advantageous to directly align the streamlines because the result
would be closely related to the final goal of registration.

Similar to~\cite{odonnell2012unbiased}, in this work, we explore the
idea of working on deterministic tractography rather than other
images. However, in contrast to all current tractography registration
methods, which are based on rigid or non-rigid shape transformation of
one tractography into another, our approach tries to find which
streamline of one tractography corresponds to which streamline in the
other tractography, without transformations. This correspondence is a
\emph{mapping} from one tractography to the other.

In this work we propose to solve the problem of finding the mapping
between two tractographies through a graph-based approach similar to
that of the well-known graph matching
problem~\cite{conte2004thirty,zaslavskiy2009path}. In the graph
matching problem the aim is to find which node of one graph
corresponds to which node of another graph, under the assumption that
graphs have the same number of nodes and that the correspondence is
one-to-one.

Given a tractography of $N$ streamlines $T = \{s_1,\ldots,s_N\}$ and a
distance function $d$ between streamlines, we can create an undirected
weighted graph by considering each streamline as a vertex and the edge
connecting vertex $s_i$ and $s_j$ as the distance between the two
streamlines, $d(s_i,s_j)$. Then, intuitively, the problem of
tractography mapping becomes very similar to that of graph matching,
but with some key differences. Firstly, the size of the two
tractographies/graphs is in general not the same. Global differences
in the anatomy of the brains, e.g. different volume, motivates this
difference. Secondly, in general there is not a one-to-one
correspondence between the streamlines/nodes but a many-to-one
correspondence. This is anatomically likely if we consider that a
given anatomical structure (\emph{tract}), e.g. the cortico-spinal
tract (CST), whose streamlines should have direct correspondence
across subjects, may have different thickness, i.e. different number
of streamlines. In this case, for example, multiple streamlines of one
CST would correspond to a single streamline in the other CST. Because
of these differences, it is generally not possible to directly apply
efficient graph matching algorithms to the problem of mapping
tractographies.

In the following we formally describe the tractography mapping problem
starting from the graph matching problem and define the details of the
optimization problem to solve. We provide a preliminary algorithmic
solution, based on simulated annealing, to minimize the proposed loss
function. Then, we apply our proposed solution to a tractography
segmentation task in order to compare a standard registration-based
method to our proposed method on a fair ground. We conclude the paper
with a brief discussion of the preliminary encouraging results.



\section{Methods}
\label{sec:methods}
An undirected weighted graph $G=(V,E)$ of size $N$ is a finite set of
vertices $V=\{1,\ldots,N\}$ and edges $E \subset V \times V$. The
graph matching problem can be described as follows. Given two graphs
$G_A$ to $G_B$ with the \emph{same} number of vertices $N$, the
problem of matching $G_A$ and $G_B$ is to find the correspondence
between vertices of $G_A$ and vertices of $G_B$, which allows to
align, or register, $G_A$ and $G_B$ in some optimal way. The
correspondence between vertices of $G_A$ and of $G_B$ is defined as a
\emph{permutation} $P$ of the $N$ vertices, i.e. there a one-to-one
correspondence between the two set of vertices. $P$ is usually
represented as a binary $N\times N$ matrix where $P_{ij}$ is equal to
$1$, if the $i$th vertex of $G_A$ is matched to the $j$th vertex of
$G_B$, otherwise $0$. Given $A$ and $B$, i.e. the $N\times N$
adjacency matrices of the two graphs, the quality of the matching is
assessed by the discrepancy, or loss, between the graphs after
matching as:
\begin{equation}
\label{eq:graph_matching_loss}
L(P) = \Vert A - P B P^\top \Vert_2 
\end{equation}
where $\Vert A \Vert_2 = \sqrt{\sum_{ij}^N A_{ij}^2}$ is the
Frobenius norm. Therefore, the graph matching problem becomes the
problem of finding $P^*$ that minimize $L$ over the set of permutation
matrices $\mathcal{P}$:
\begin{equation}
  \label{eq:graph_matching}
  P^* = \argmin_{P \in \mathcal{P}} \Vert A - P B P^\top \Vert_2 
\end{equation}
which is a combinatorial optimization problem. The exact solution to
this problem is NP-complete and only approximate solutions are
available in practical
cases~\cite{conte2004thirty,zaslavskiy2009path}.

Let $T_A = \{s^A_1,\ldots,s^A_N\}$ and $T_B = \{s^B_1,\ldots,s^B_M\}$,
where $s = \{x_1,\ldots,x_{n_s}\}$ is a streamline and $x \in
\mathbb{R}^3$, be the tractographies of two subjects. Let $d$ be a
distance function between streamlines. We define two graphs $G_A$ and
$G_B$ with adjacency matrix $A \in \mathbb{R}^{N \times N}$ and $B \in
\mathbb{R}^{M \times M}$ where $A_{ij} = d(s^A_i,s^A_j)$ and $B_{ij} =
d(s^B_i,s^B_j)$. Our current choice of $d$ is discussed in
Section~\ref{sec:experiments}, however any common streamline distance
from the literature can be used.

The loss function of a \emph{mapping} $Q$ from $T_A$ to $T_B$ is then:
\begin{equation}
\label{eq:mapping_loss}
L(Q) = \Vert A - Q B Q^\top \Vert_2 
\end{equation}
where the mapping $Q$ is a binary $N\times M$ matrix and $Q_{ij}$ is
equal to $1$, if $s_i^A$ of $T_A$ is mapped to $s_j^B$ of $T_B$ and
$0$ otherwise. Note that, in general, $Q$ is not a permutation matrix,
because multiple streamlines can be mapped into the same one. In order
to find the optimal mapping $Q^*$, we minimize $L$ so that $T_B$ is
most similar to $T_A$:
\begin{equation}
  \label{eq:mapping}
  Q^* = \argmin_{Q \in \mathcal{Q}} \Vert A - Q B Q^\top \Vert_2
\end{equation}
where $\mathcal{Q}$ is the set of all possible mappings. Because in
general $N \neq M$ and because $Q$ is a mapping and not just a
permutation, the tractography mapping problem has a larger search
space than the graph matching problem, i.e. $|\mathcal{Q}| = M^N \gg
N!  = |\mathcal{P}|$ when $M \approx N$, is much larger than
$\mathcal{P}$. As a consequence, the efficient solutions available in
the literature of graph matching, e.g.~\cite{zaslavskiy2009path}, are
not applicable, because they heavily rely on the assumptions that we
violate here. In Section~\ref{sec:experiments} we implemented a simple
preliminary solution to the combinatorial optimization problem by
means of the Simulated Annealing
meta-heuristic~\cite{laarhoven1987simulated}.


\subsection{Comparison}
In order to compare the proposed method against a standard
registration procedure on a fair ground, we cannot rely on the value
of the loss function $L$, because it is defined only in the case of
mapping. For this reason, we compared the two approaches on the
practical task of automatic tractography segmentation, i.e. finding a
given tract of interest in $T_B$ given its segmentation in $T_A$. Our
hypothesis is that reducing $L$ leads to better overlap between
tractographies, which is important for practical applications like
segmentation. In Section~\ref{sec:experiments} we describe an
experiment to test this hypothesis and provide the necessary
details. Here we introduce the metric that we use for comparing
registration and mapping. As proposed in~\cite{golding2011comparison},
we compare the set of voxels crossed by the streamlines of each
tractography after mapping or after registration. As measure of the
overlap between $T_A$ and $Q(T_B)$\footnote{For sake of brevity we
  denote as $Q(T_B)$ the result of applying mapping $Q$ to $T_B$.}, we
adopt the Jaccard index:
\begin{equation}
  \label{eq:Jaccard}
  J(T_A, T_B|Q) = \frac{|T_{A} \cap Q(T_B)|}{ \min\{|T_A|, |Q(T_B)|\}}
\end{equation}
Note that in the above equation, $|T|$ is the volume computed as
number of voxels that any streamline $s \in T$ goes through, and
$|T_A \cap Q(T_B)|$ indicates the number of voxels in common between
$T_A$ and $Q(T_B)$.


\section{Experiments}%
\label{sec:experiments}
We designed an experiment to provide empirical evidence that reducing
the loss in Equation~\ref{eq:mapping_loss} is related to an increase
of the Jaccard index, i.e. of the overlap between tractographies.

The dataset used for the experiment is based on dMRI data recorded
with a $3T$ scanner at Utah Brain Institute, $65$ gradients ($64$ +
$b0$); b-value = $1000$; anatomical scan ($2\times2\times2mm^3$). The
tractography was reconstructed with the EuDX
algorithm~\cite{garyfallidis2012towards} using the
dipy\footnote{\url{http://www.dipy.org}} toolbox. We considered 4
healthy subjects and focused the analysis on the corticospinal tract
(CST). CST is a set of streamlines projecting from the lateral medial
cortex associated with the motor homunculus. This tract is of main
interest for the characterization of neurodegenerative diseases, like
the amyotrophic lateral sclerosis (ALS). The CST tracts were segmented
by the expert neuroanatomists using a toolbox~\cite{olivetti2013fast}
that supports an interactive selection of streamlines. The size of the
segmented tracts is reported in Table~\ref{table:JAC_205_left} (see
column \emph{size}).

The reference method, against which we compared mapping, is the affine
registration of the tractographies in a common MNI space using the
voxel-based FLIRT method~\cite{jenkinson2001global}. The registration
is defined as follows: First, FA images were registered to the
MNI-FMRIB-58 FA template, then the affine transformation was applied
to the tractographies. The Jaccard index computed between the $CST_A$
and $CST_B$ in common space is reported in
Table~\ref{table:JAC_205_left} (see column FLIRT).

We then used mapping to compute the same quantity. The first step was
encoding the tractographies as graphs, which required to define a
distance between streamlines. We refer to the commonly used Mean
Average Minimum distance (MAM)~\cite{zhang2008identifying}, based on
the Hausdorff distance:
\begin{equation}
  \label{equ:mam}
  d_{MAM}(s,s') = \frac{1}{2} (D(s,s') + D(s',s))
\end{equation}
where $D(s,s') = \frac{1}{n_s} \sum_{i=1}^{n_s} d(x_i,s')$, and $d(x, s') =
\min_{j = 1,\ldots,n_{s'}}||x-x'_j||_2$.

Mapping a tract such as the CST, which usually comprises $10^2$
streamlines, to an entire tractography $T_B$, which usually consist of
$10^7$ streamlines, is computationally extremely expensive because the
space of all possible mappings $\mathcal{Q}$ has size
$|T_B|^{|CST|}$. For this reason, we introduced a heuristic to retain
some of the streamlines in $T_B$. The intuitive idea was to define a
superset of streamlines of the CST for subject B, denoted
$CST_B^+$. The heuristic is in two steps: first, we computed the
medoid $s_m$ of $CST_B$, and the radius $r = \max\{d(s_m,s_i), \forall
s_i \in CST_B\}$. Second, we filtered the streamlines in $T_B$ such
that $CST_{B}^+ = \{s_j \in T_B | d(s_{m}, s_j) \leq \alpha \cdot r\}
$, where $\alpha = 3$. See Table~\ref{table:JAC_205_left}, column
$CST_B⁺$, for the actual sizes of the supersets.


Computing the optimal mapping $Q^*$ requires to solve, even in an
approximate way, the minimization problem of
Equation~\ref{eq:mapping}. As a preliminary strategy to approximate
the optimal mapping $Q^*$, we implemented the simulated annealing
(SA)~\cite{laarhoven1987simulated} meta-heuristic, a reference method
for combinatorial optimization. SA requires the definition of a
function to move from the current state, i.e. the current mapping $Q$,
to a (potentially better) neighbouring one. As transition function we
used a stochastic greedy one where, given the current mapping $Q$, one
streamline of $CST_A$ is selected at random and then it is greedily
re-mapped to the streamline in $CST_B^+$ providing the greatest
reduction in the loss of Equation~\ref{eq:mapping_loss}. As starting
point of the annealing process, we used the $1$-nearest neighbour of
$CST_A$ with respect to $CST_{B}^+$ after the registration of $T_A$
and $T_B$. We ran the simulated annealing for $1000$ iterations, which
required a few minutes on a standard computer\footnote{We are aware
  that this method of combinatorial optimization can be significantly
  improved, but we claim that the it was sufficient to do a
  preliminary investigation of the relation between the loss $L$ and
  the overlap between tractographies, by means of the Jaccard index.}.

\begin{figure}[t]
\centering   
  \includegraphics[width=8cm ]{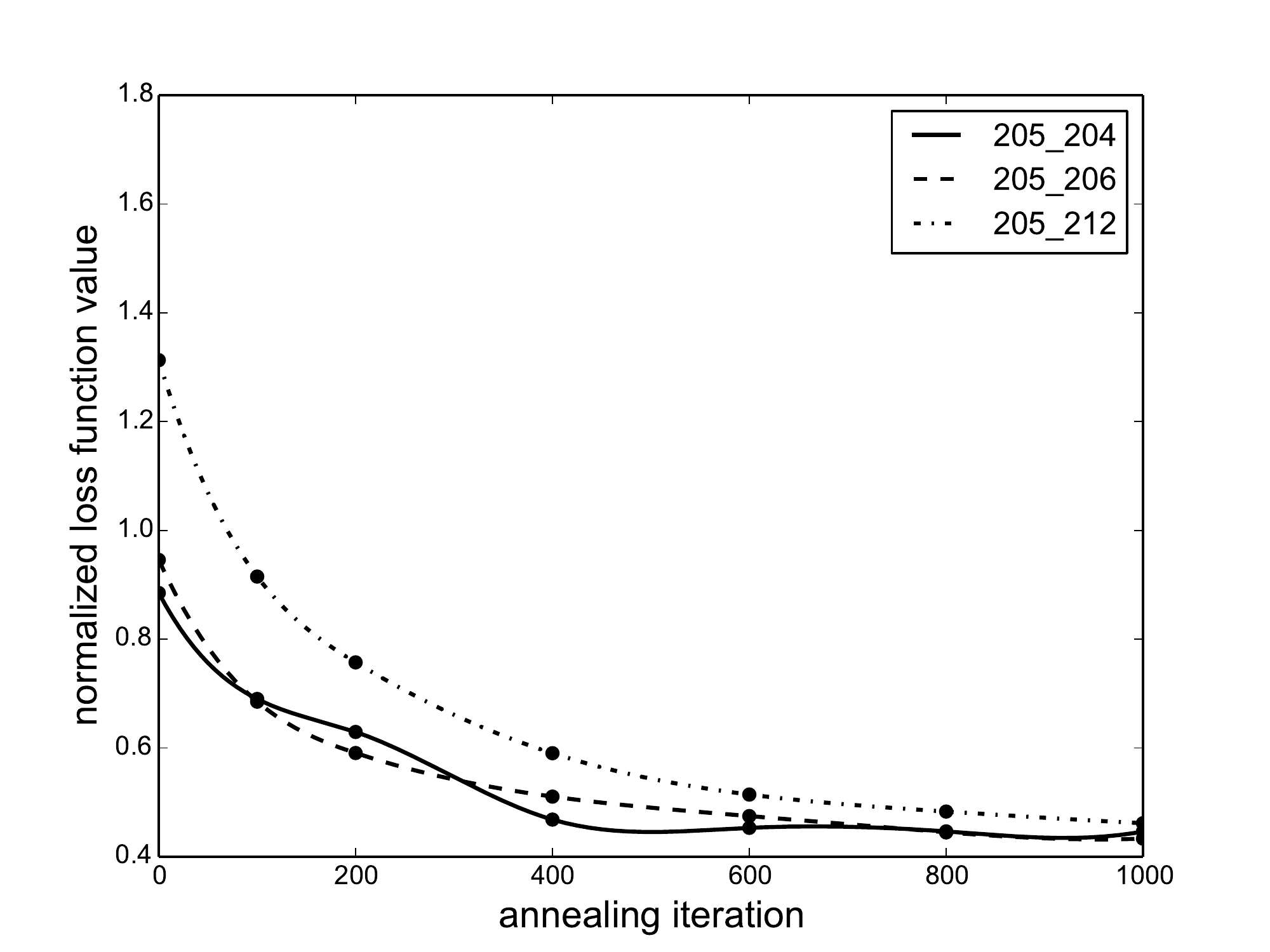}  
  \caption{Plots of the normalized loss ($L_{norm} = \frac{L}{|CST_A|}
    $) as a function of number of iterations with simulated annealing,
    when mapping the CST of subject 205 to those of subjects 204, 206
    and 212.}
  \label{fig:loss_anneal}
\end{figure}

\begin{figure}[hbt]
  \centering
  \includegraphics[width=12cm]{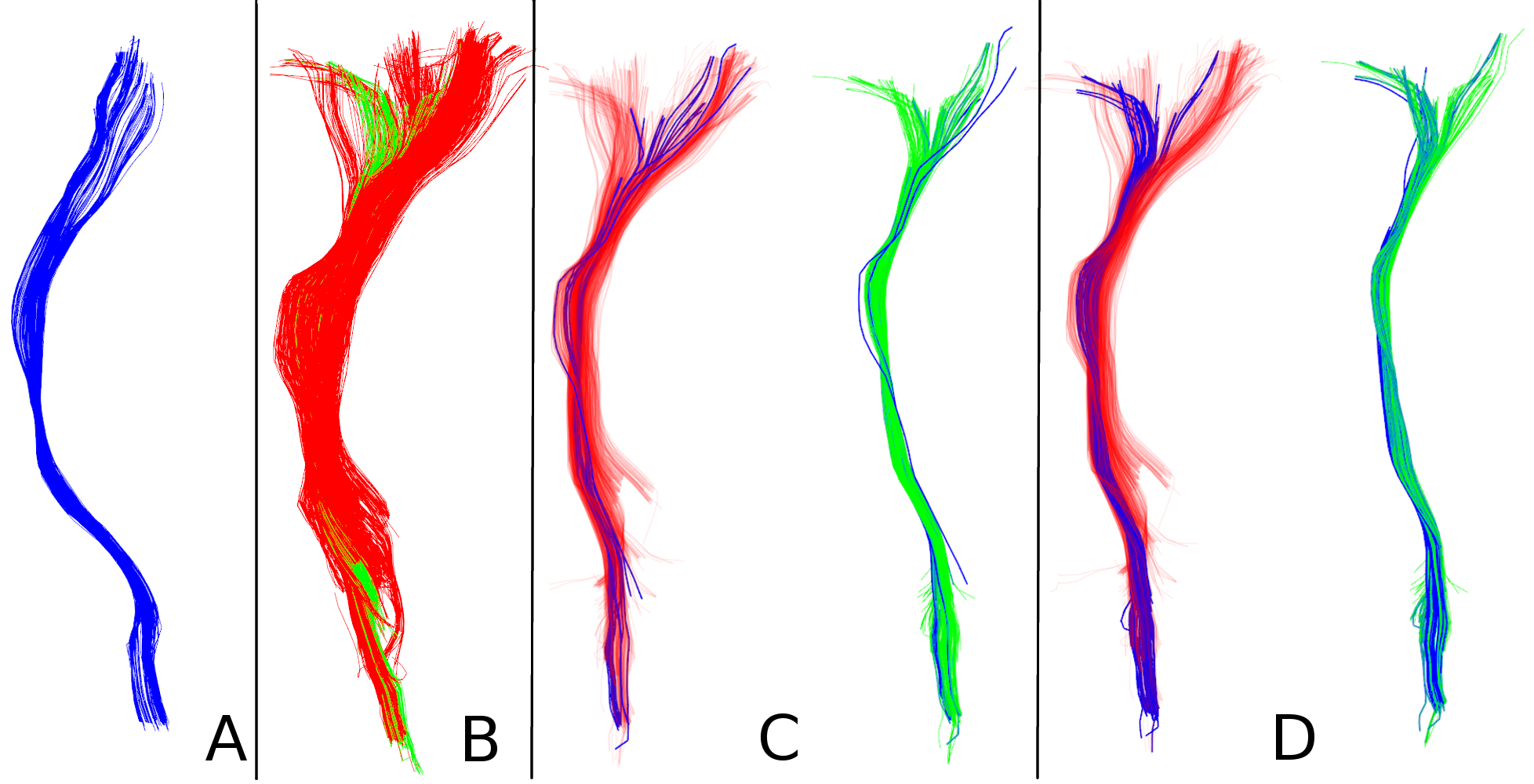}
  \caption{Example of registration vs. mapping of the corticospinal
    tract (CST). From the left, the tract to be mapped (subfigure A,
    $CST_A$ in blue), the second tract with its superset (subfigure B,
    $CST_B$ in green, $CST^+_B$ in red), the result of FLIRT affine
    registration (subfigure C) and of mapping (subfigure D).}
  \label{fig:example_comparison}
\end{figure}

The results reported in Figure~\ref{fig:loss_anneal} show the
behaviour of the loss during the optimization process for the mapping
of $CST_A$ (subject ID $205$), with respect to the tractography of
three other subjects (subject IDs $204$, $206$ and $212$). In all
cases, as the number of iterations increases, the value of loss
function decreases. In Figure~\ref{fig:example_comparison} we show an
example of experiment with the outcome of FLIRT registration and
mapping which refers to subjects 204 and 206. In subfigure A, the
source tract $CST_A$ is shown in blue, in subfigure B the target tract
$CST_B$ is show in green and the related superset of streamlines
$CST_B^+$ in red. In subfigure C, the result of FLIRT registration is
presented, both with respect to the superset $CST_B^+$ on the left and
with respect to the target tract $CST_B$ on the right. On the right
side, it is illustrated the set of streamlines (blue) from the source
tract $CST_A$ associated to streamlines of target tract $CST_B$. The
association between streamlines of $CST_A$ and $CST_B$ is computed as
nearest neighbour after the FLIRT registration. The ratio between blue
and green streamlines represents the portion of target tract correctly
detected. On the left side of subfigure C, blue streamlines represents
the portion of source tract $CST_A$ not associated to target tract
$CST_B$. In subfigure D, the result of mapping is presented, with the
same strategy of presentation of subfigure C. On the right side the
visualization shows a greater amount of (blue) streamlines correctly
mapped into target tract. Even on the left side the amount of (blue)
streamlines erroneously mapped is greater. The sum of blue streamlines
on the left and right side represents the portion of streamlines
projected from the source to the target. The registration based on
FLIRT doesn't preserve after the alignment the same amount of
streamlines from the source tract.

\begin{table}[hbt]
\centering 
\setlength{\tabcolsep}{3.pt}
\renewcommand{\arraystretch}{1.28} 
\begin{tabular}[c]{| c | c | c | c | c | c | c | c | c |}
\hline
A & B & \multicolumn{3}{c|}{size} &  \multicolumn{4}{c|}{ Jaccard index}\\
\cline{3-9}
subject ID &  subject ID & \it $|CST_A|$  & $|CST_B|$  & $|CST_{B}^+|$ & \it FLIRT & \it SA-0 & \it SA-100 & \it SA-1000 \\
\hline
205	&	204	&	60	&	124	& 	682	&	0.18 &	0.55 &	0.52 &	\textbf{0.59}\\
	&	206	&	60	&	100 &	550 &	0.15 &	0.77 &	0.81 &	\textbf{0.82}\\
	&	212	&	60	&	68	&	374	&	0.10 &	0.74 &	0.77 &	\textbf{0.90}\\
\hline
\end{tabular}
\caption{Comparison of registration vs. mapping. The subject IDs of
  $CST_A$ and $CST_B$ are reported in the first two columns. Their
  sizes, together with that of $CST_B^+$, are in columns three to
  five. The last four columns report the overlap between
  $CST_A$ and $CST_B$ in terms of Jaccard index (higher is better),
  for FLIRT registration (6th column) and for mapping with simulated
  annealing at a different number of iterations (SA-0, SA-100, SA-1000
  columns).}
\label{table:JAC_205_left}
\end{table}
 
In Table~\ref{table:JAC_205_left} are reported the results of the
comparison between registration and mapping methods, measured by the
Jaccard index. The overlap between $CST_A$ and $CST_B$ provided by
FLIRT registration is generally quite poor. This is partly expected
because even after the registration of $T_A$ and $T_B$, $CST_A$ and
$CST_B$ may have a systematic displacement due to the variability of
anatomy across subjects. The results of mapping at different
iterations of the optimization process shows a remarkable global
increase in the Jaccard index and a general trend of improved
alignment when more iterations are computed.


\section{Discussion and Conclusion}%
\label{sec:discussion}

In this work we addressed the challenge of finding an alignment
between the tractographies of two subjects. We recast the question as
a problem of mapping between two sets of streamlines and we provided
the formulation of the corresponding minimization problem. Preliminary
results show that this approach is promising despite some limitations.
The computational complexity represents a major issue that may prevent
to scale up to whole tractography.





\end{document}